\documentclass[letterpaper]{article}
\usepackage[submission]{aaai24}
\usepackage{times}
\usepackage{helvet}
\usepackage{courier}
\usepackage[hyphens]{url}
\usepackage{graphicx}
\usepackage{natbib}
\usepackage{caption}
\usepackage{algorithm}
\usepackage[noend]{algorithmic}

\usepackage{newfloat}
\usepackage{listings}
\DeclareCaptionStyle{ruled}{labelfont=normalfont,labelsep=colon,strut=off}
\floatstyle{ruled}
\newfloat{listing}{tb}{lst}{}
\floatname{listing}{Listing}
\usepackage{amsmath}
\usepackage{amsfonts}
\usepackage{todonotes}
\usepackage{booktabs}
\usepackage{subcaption}
\usepackage{amsthm}
\usepackage{adjustbox}
\usepackage{multirow}
\usepackage{nicematrix}

\usepackage{tikz}
\usetikzlibrary{positioning, arrows, arrows.meta, trees, shapes.geometric}

\newcommand{\pbcount}{\ensuremath{\mathsf{PBCount}}}
\newcommand{\apply}{\ensuremath{\mathsf{Apply}}}
\newcommand{\ite}{\ensuremath{\mathsf{ITE}}}

\newcommand{\func}[1]{\ensuremath{\mathsf{Func}}(#1)}
\newcommand{\dpmc}{\ensuremath{\mathsf{DPMC}}}
\newcommand{\gpmc}{\ensuremath{\mathsf{GPMC}}}
\newcommand{\dfour}{\ensuremath{\mathsf{D4}}}
\newcommand{\addmc}{\ensuremath{\mathsf{ADDMC}}}
\newcommand{\pblib}{\ensuremath{\mathsf{PBLib}}}
\newcommand{\pbencoder}{\ensuremath{\mathsf{PBEncoder}}}
\newcommand{\roundingsat}{\ensuremath{\mathsf{RoundingSat}}}

\title{Engineering an Exact Pseudo-Boolean Model Counter\thanks{The code is available at https://github.com/grab/pbcount}}
\author {
    Suwei Yang\textsuperscript{\rm 1,2,3},
    Kuldeep S. Meel\textsuperscript{\rm 3,4}
}
\affiliations {
    \textsuperscript{\rm 1}GrabTaxi Holdings\\
    \textsuperscript{\rm 2}Grab-NUS AI Lab\\
    \textsuperscript{\rm 3}National University of Singapore\\
    \textsuperscript{\rm 4}University of Toronto\\
}

\begin{document}

\maketitle

\begin{abstract}
Model counting, a fundamental task in computer science, involves determining the number of satisfying assignments to a Boolean formula, typically represented in conjunctive normal form (CNF). While model counting for CNF formulas has received extensive attention with a broad range of applications, the study of model counting for Pseudo-Boolean (PB) formulas has been relatively overlooked. Pseudo-Boolean formulas, being more succinct than propositional Boolean formulas, offer greater flexibility in representing real-world problems. Consequently, there is a crucial need to investigate efficient techniques for model counting for PB formulas.

In this work, we propose the first exact Pseudo-Boolean model counter, {\pbcount}, that relies on knowledge compilation approach via algebraic decision diagrams. Our extensive empirical evaluation shows that {\pbcount} can compute counts for 1513 instances while the current state-of-the-art approach could only handle 1013 instances. Our work opens up several avenues for future work in the context of model counting for PB formulas, such as the development of preprocessing techniques and exploration of approaches other than knowledge compilation. 
\end{abstract}

\section{Introduction} \label{sec:introduction}

Propositional model counting involves computing the number of satisfying assignments to a Boolean formula. Model counting is closely related to the Boolean satisfiability problem where the task is to determine if there exists an assignment of variables such that the Boolean formula evaluates to \textit{true}. Boolean satisfiability and model counting have been extensively studied in the past decades and are the cornerstone of an extensive range of real-life applications such as software design, explainable machine learning, planning, and probabilistic reasoning~\cite{BDP03,NSMIM19,J19,FMM20}. Owing to decades of research, there are numerous tools and techniques developed for various aspects of Boolean satisfiability and model counting, from Boolean formula preprocessors to SAT solvers and model counters. 

The dominant representation format of Boolean formulas is Conjunctive Normal Form (CNF), and accordingly, the tools in the early days focused on CNF as the input format. Over the past decade and a half, there has been considerable effort in exploring other representation formats: one such format that has gained significant interest from the community is Pseudo-Boolean (PB) formulas, which are expressed as the conjunction of linear inequalities. PB formulas are shown to be more succinct than CNF formulas and natural for problems such as Knapsack, sensor placement, binarized neural networks, and the like. Furthermore, PB formulas are able to express constraints more succinctly compared to Boolean formulas in CNF~\cite{LMMW18}. As an example, a single PB constraint is sufficient to express at-most-$k$ and at-least-$k$ types of cardinal constraints whereas the equivalent in CNF would require a polynomial number of clauses~\cite{S05}. On a higher level, an arbitrary CNF clause can be expressed with a single PB constraint but the converse is not true~\cite{LMMW18}. The past decade has witnessed the development of satisfiability solving techniques based on the underlying proof systems naturally suited to PB constraints, and accordingly, the state-of-the-art PB solvers, such as {\roundingsat} significantly outperform CNF solvers on problems that are naturally encoded in PB~\cite{EN18,D20,DGDNS21}. 

In contrast to satisfiability, almost all the work in the context of model counting has focused on the representation of Boolean formulas in Conjunctive Normal Form (CNF), with the sole exception of the development of an approximate model counter for PB formulas~\cite{YM21}. 

The primary contribution of this work is to address the aforementioned gap through the development of a native scalable exact model counter, called {\pbcount}, for PB formulas. {\pbcount} is based on the knowledge compilation paradigm, and in particular, compiles a given PB formula into algebraic decision diagrams (ADDs)~\cite{BFGHMPF93}, which allows us to perform model counting. We perform extensive empirical evaluations on benchmark instances arising from different applications, such as sensor placement, multi-dimension knapsack, and combinatorial auction benchmarks~\cite{GL80,BN07,LSM23}. Our evaluations highlighted the efficacy of {\pbcount} against existing state-of-the-art CNF model counters. In particular, {\pbcount} is able to successfully count 1513 instances while the prior state of the art could only count 1013 instances, thereby demonstrating significant runtime improvements. It is worth remarking that {\pbcount} achieves superior performance with substantially weaker preprocessing techniques in comparison to techniques employed in CNF model counters, making a strong case for the advantages of native PB model counting and reasoning. Furthermore, given the crucial importance of preprocessing techniques for CNF counting, we hope our work will motivate the development of preprocessing techniques for PB model counting.

The rest of the paper is organized as follows: We discuss the preliminaries and existing counting algorithm in Section~\ref{sec:background}. In Section~\ref{sec:related-works}, we discuss existing works and how they relate to our approach, which we detail in Section~\ref{sec:approach}. Following that, we analyze the empirical results of {\pbcount} against existing tools in Section~\ref{sec:experiments} and conclude in Section~\ref{sec:conclusion}.

\section{Preliminaries} \label{sec:background}

\paragraph{Boolean Formula}

A Boolean variable can take values \textit{true} or \textit{false}. A literal is either a Boolean variable or its negation. Let $F$ be a Boolean formula. $F$ is in conjunctive normal form (CNF) if $F$ is a conjunction of clauses, where each clause is a disjunction of literals. $F$ is satisfiable if there exists an assignment $\tau$ of variables of $F$ such that $F$ evaluates to \textit{true}. We refer to $\tau$ as a satisfying assignment of $F$ and denote the set of all $\tau$ as $\mathsf{Sol}(F)$. Model counting for Boolean formula $F$ refers to the task of determining $|\mathsf{Sol}(F)|$.

\paragraph{Pseudo-Boolean Formula}

A PB constraint is either an equality or inequality of the form $\sum_{i=1}^{n} a_i x_i \square k$ where $x_1, ..., x_n$ are Boolean literals, $a_1, ..., a_n$, and $k$ are integers, and $\square$ is one of $\{\geq, =, \leq\}$. We refer to $a_1, ..., a_n$ as term coefficients in the PB constraint, where each term is of the form $a_i x_i$. A PB formula, $G$, consists of a set of PB constraints. $G$ is satisfiable if there exists an assignment $\tau$ of all variables of $G$ such that all its PB constraints hold. PB model counting refers to the computation of $|\mathsf{Sol}(G)|$ where $\mathsf{Sol}(G)$ is the set of all satisfying assignments of $G$.

\paragraph{Projected Model Counting}
Let $G$ be a formula defined over the set of variables $X$. Let $V_i, V_j$ be subsets of $X$ such that $V_i \cap V_j = \emptyset$  and $V_i \cup V_j = X$. Projected model counting of $G$ on $V_i$ refers to the number of assignments of all variables in $V_i$ such that there exists an assignment of variables in $V_j$ that makes $G$ evaluate to \textit{true}~\cite{ACMS15}. In the evaluations, CNF model counter baselines perform projected model counting on the original variables in the PB formula, to avoid additional counts due to auxiliary variables introduced in the PB to CNF conversion process.

\paragraph{Algebraic Decision Diagram}

\begin{figure}[htb]
    \centering
    \begin{adjustbox}{width=0.45\columnwidth}
    \begin{tikzpicture}[
    roundnode/.style={circle, draw=black!60, very thick, minimum size=7mm},
    ]
    \node[roundnode](x1) at (0, 0){$x_1$};
    \node[roundnode](x2left) at (-1, -1.25){$x_2$};
    \node[roundnode](x2right) at (1, -1.25){$x_2$};
    \node[roundnode](leaf1) at (-2, -2.75){$0$};
    \node[roundnode](leaf2) at (-0.75, -2.75){$4$};
    \node[roundnode](leaf3) at (0.75, -2.75){$3$};
    \node[roundnode](leaf4) at (2, -2.75){$7$};

    \draw[dashed,->] (x1) -- (x2left);
    \draw[->] (x1) -- (x2right);
    \draw[dashed,->] (x2left) -- (leaf1);
    \draw[->] (x2left) -- (leaf2);
    \draw[dashed,->] (x2right) -- (leaf3);
    \draw[->] (x2right) -- (leaf4);
	
    \end{tikzpicture}
    \end{adjustbox}
    \caption{An ADD representing $3x_1 + 4x_2$}
    \label{fig:add-expression}
\end{figure}
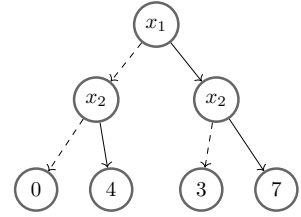

An algebraic decision diagram (ADD) is a directed acyclic graph representation of a function $f: 2^{X} \rightarrow S$ where $X$ is the set of Boolean variables that $f$ is defined over, and $S$ is an arbitrary set known as the carrier set. We denote the function represented by an ADD $\psi$ as $\func{\psi}$. The internal nodes of ADD represent decisions on variables $x \in X$ and the leaf nodes represent $s \in S$. In this work, we focus on the setting where $S \subset \mathbb{Z}$. As an example, an ADD representing $3x_1 + 4x_2$ is shown in Figure~\ref{fig:add-expression}. In the figure, a dotted arrow from an internal node represents when the corresponding variable is set to \textit{false} and a solid arrow represents when it is set to \textit{true}.

In addition, we make use of {\apply} and {\ite} operations on ADDs~\cite{B86,BFGHMPF93}. The {\apply} operation takes as input a binary operator $\bowtie$, two ADDs $\psi_1, \psi_2$, and outputs an ADD $\psi_3$ such that the $\func{\psi_3} = \func{\psi_1} \bowtie \func{\psi_2}$. The {\ite} operation (\textit{if-then-else}) involves 3 ADDs $\psi_1, \psi_2, \psi_3$, where carrier set of $\psi_1$ is restricted to $\{0,1\}$. {\ite} outputs an ADD that is equivalent to having 1 valued leaf nodes in $\psi_1$ replaced with $\psi_2$ and 0 valued leaf nodes with $\psi_3$.

\paragraph{Relation of Pseudo-Boolean Constraint to CNF Clause}

Given an arbitrary CNF clause $D$, one could always convert $D$ to a PB constraint. Given that $D$ is of the form $\bigvee_{i=1}^{m} l_i$, where $l_1, ... , l_m$ are Boolean literals, $D$ can be represented by a single PB constraint $\sum_{i=1}^{m} a_i l_i \geq 1$ where all coefficients $a_1, ..., a_i, ...a_m$ are $1$. However, there are PB constraints that require polynomially many CNF clauses to represent. An example would be $\sum_{i=1}^{m} l_i \geq k$ which requires at least $k$ of $m$ literals to be \textit{true}. We refer the reader to the Appendix for statistics of the number of variables and clauses before and after PB to CNF conversion for benchmarks used.

\subsection{Model Counting with ADDs} \label{subsec:add-model-count}
In this work, we adapt the existing dynamic programming counting algorithm of {\addmc}~\cite{DPV20a}, shown in Algorithm~\ref{alg:add-count}, to perform PB model counting with ADDs. This includes using the default {\addmc} configurations for ADD variable ordering (MCS) and cluster ordering $\rho$ (BOUQUET\_TREE). The algorithm takes in a list $\varphi$ of ADDs, representing all constraints, and an order $\rho$ of which to process the ADDs. ADD $\psi$ is initialized with value $1$. According to cluster ordering $\rho$, cluster ADDs $\psi_j$ are formed using the {\apply} operation with $\times$ operator on each of the individual constraint ADDs of constraints in the cluster. The cluster ADD $\psi_j$ is combined with $\psi$ using the same {\apply} operation. If variable $x$ does not appear in later clusters in $\rho$, it is abstracted out from $\psi$ (early projection process in {\addmc}) using $\psi \gets \mathcal{W}(\bar{x}) \times \psi[x \mapsto 0] + \mathcal{W}(x) \times \psi[x \mapsto 1]$ in line~\ref{line:add-count:early-projection}, where $\mathcal{W}(\cdot)$ is the user provided literal weight function. In unweighted model counting, $\mathcal{W}(\cdot)$ is 1 for all literals. Once all clusters have been processed, the unprocessed variables $x$ of the formula $G$ are abstracted out using the same operation as before (line~\ref{line:add-count:remaining-projection}). After all variables are abstracted out, $\psi$ is a constant ADD that represents the final count.

\begin{algorithm}
\caption{computeCount($\varphi, \rho$)} \label{alg:add-count}
\textbf{Input:} $\varphi$ - list of ADD, $\rho$ - cluster merge ordering

\textbf{Output:} model count
\begin{algorithmic}[1]
    \STATE $\psi \gets \mathsf{constantADD}(1)$
    \FOR{$\mathsf{cluster}$ $A_j \in \rho$}
        \STATE $\psi_j \gets \mathsf{constantADD}(1)$
        \FOR{$\mathsf{constraint}$ $C_i \in A_j$}
            \STATE $\psi_j \gets \psi_j \times \varphi[C_i]$
        \ENDFOR
        \STATE $\psi \gets \psi \times \psi_j$
        \FOR{each $x \in \psi$ where $x$ not in later clusters in $\rho$}
            \STATE $\psi \gets \mathcal{W}(\bar{x}) \times \psi[x \mapsto 0] + \mathcal{W}(x) \times \psi[x \mapsto 1]$ \label{line:add-count:early-projection}
        \ENDFOR
    \ENDFOR
    \FOR{all unprocessed variable $x$}
        \STATE $\psi \gets \mathcal{W}(\bar{x}) \times \psi[x \mapsto 0] + \mathcal{W}(x) \times \psi[x \mapsto 1]$ \label{line:add-count:remaining-projection}
    \ENDFOR
    \RETURN $\mathsf{getValue}(\psi)$
\end{algorithmic}
\end{algorithm}

\section{Related Work} \label{sec:related-works}

\paragraph{Boolean Formula Preprocessing}

Boolean formula preprocessing involves simplifying a given formula to reduce runtimes of downstream tasks such as determining satisfiability of the formula (SAT-solving) and model counting. Preprocessing is crucial to modern SAT solvers and model counters' performance improvements in recent decades. There are numerous preprocessing techniques introduced over the years by the research community, some of which are \textit{unit propagation}, \textit{bounded variable elimination}, \textit{failed literal probing}, and \textit{vivification}~\cite{DG84,L01,EB05,PHS08}. In this work, we adapt some of the SAT preprocessing techniques, namely \textit{unit propagation} and a variant of \textit{failed literal probing}, to simplify PB formulas. 

\paragraph{Search-Based Model Counters}
Among the numerous existing CNF model counters, we can classify them into two main categories -- search-based model counters and decision diagram-based model counters. Notable existing search-based model counters include {\gpmc}, $\mathsf{Ganak}$, and \textsf{Sharpsat-TD}~\cite{SHS17,SRSM19,KJ21}. Search-based model counters work by setting values to variables in a given formula in an iterative manner, which is equivalent to implicitly exploring a search tree. In addition, search-based model counters adapt techniques such as sub-component caching from SAT solving for more efficient computation.

\paragraph{Decision Diagram-Based Model Counter}
Decision diagram-based model counters employ knowledge compilation techniques to compile a given formula into directed acyclic graphs (DAGs) and perform model counting with these DAGs. Some of the recent decision diagram-based model counters are {\dfour}, \textsf{ExactMC}, {\addmc}, and its related variant {\dpmc}~\cite{LM17,DPV20a,DPV20b,LMY21}. {\dfour} and \textsf{ExactMC} compile the formula in a top-down manner into the respective decision diagram forms. In contrast, {\addmc} and {\dpmc} (decision diagram mode) perform bottom-up compilations of algebraic decision diagrams (ADDs). In this work, we based {\pbcount} on {\addmc} and introduced techniques to compile a PB constraint directly into an ADD and employ the same counting approach in {\addmc}.

\paragraph{Pseudo-Boolean Conversion}
One way to perform PB model counting is to convert the PB formula to a Boolean formula and use existing CNF model counters. A notable tool for the conversion of PB to CNF is {\pblib}~\cite{PS15}. {\pblib} implements various encodings to convert PB formulas into CNF form, some of which include cardinality networks, sorting networks, and BDD-based encodings~\cite{ES06,ANOR11,ANOR13}. In this work, we use default settings for the {\pbencoder} binary provided as part of {\pblib} to perform the required conversions. We subsequently compare {\pbcount} against state-of-the-art CNF model counters. It is worth noting that the model counting task for PB formula becomes a projected model counting task of the corresponding CNF formula, as previously mentioned in Section~\ref{sec:background}.

\section{Approach} \label{sec:approach}

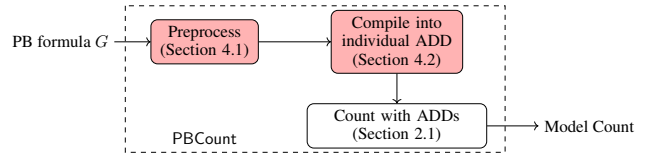
\begin{figure}[htb]
    \centering
    \begin{adjustbox}{width=1.0\columnwidth}
    \begin{tikzpicture}[
    node distance=2cm,
    diamondnode/.style={diamond, minimum width=1.5cm, minimum height=0.5cm, text centered, draw=black, fill=green!30},
    rectnode/.style={rectangle, rounded corners, minimum width=1.5cm, minimum height=0.5cm,text centered, draw=black, fill=red!30},
    rectnodenofill/.style={rectangle, rounded corners, minimum width=1.5cm, minimum height=0.5cm,text centered, draw=black},
    ]
    \node (preprocess) [rectnode, text width=2.0cm] at (0,0) {Preprocess \\ (Section~\ref{subsec:preprocessing})};
    \node (compile) [rectnode, text width=2.5cm] at (4, 0) {Compile into individual ADD \\ (Section~\ref{subsec:pb-clause-compile})};

    \node (merge-count) [rectnodenofill, text width=3.5cm] at (4, -1.75) {Count with ADDs \\ (Section~\ref{subsec:add-model-count})};

    \node (pbformula) at (-3, 0) {PB formula $G$};
    \node (output-count) at (8.0, -1.75) {Model Count};

    \node (approach name) at (0, -2) {${\pbcount}$};

    \draw [->] (pbformula) -- (preprocess);
    \draw [->] (preprocess) -- (compile);
    \draw [->] (compile) -- (merge-count);
    \draw [->] (merge-count) -- (output-count);

    \draw[dashed] (-1.65, 0.75) rectangle (6.25,-2.3);

    \end{tikzpicture}
    \end{adjustbox}

    \caption{Overall flow of our PB model counter {\pbcount}. Shaded boxes indicate our contributions.}
    \label{fig:overall-flow}
\end{figure}

We show the overall flow of {\pbcount} in Figure~\ref{fig:overall-flow}. We first preprocess the PB formula using \textit{propagation} and \textit{assumption probing}. Subsequently, we compile each of the PB constraints into an algebraic decision diagram (ADD). Next, we merge constraint ADDs using {\apply} operation and perform model counting by abstracting out variables (Section~\ref{subsec:add-model-count}). 
The model count would be the value after all variables are abstracted out. Without loss of generality, the algorithms described in this work handle PB constraints involving `$=$' and `$\geq$' operators, as `$\leq$' type constraints can be manipulated into `$\geq$' type constraints.

\subsection{Preprocessing} \label{subsec:preprocessing}

\begin{figure}[htb]
    \centering
    \begin{adjustbox}{width=1.0\columnwidth}
    \begin{tikzpicture}[
    node distance=2cm,
    diamondnode/.style={diamond, minimum width=1.5cm, minimum height=0.5cm, text centered, draw=black, fill=green!30},
    rectnode/.style={rectangle, rounded corners, minimum width=1.5cm, minimum height=0.5cm,text centered, draw=black, fill=red!30},
    rectnodenofill/.style={rectangle, rounded corners, minimum width=1.5cm, minimum height=0.5cm,text centered, draw=black},
    ]
    \node (propagation) [rectnodenofill, text width=2.0cm] at (0,0) {Propagate};
    \node (probe) [rectnodenofill, text width=3.25cm] at (4, 0) {Assumption Probing};

    \node (pbformula) at (-3, 0) {PB formula $G$};
    \node (output) at (8.0, 0) {PB formula $G'$};

    \node (approach name) at (0, -0.8) {Preprocessing};

    \draw [->] (pbformula) -- (propagation);
    \draw [->] (probe) -- (output);
    \draw [->] (propagation) to [out=30,in=160] (probe);
    \draw [<-] (propagation) to [out=-30,in=-160] (probe);

    \draw[dashed] (-1.75, 0.75) rectangle (6.25,-1);

    \end{tikzpicture}
    \end{adjustbox}

    \caption{Preprocessing of PB formula}
    \label{fig:preprocessing-flow}
\end{figure}
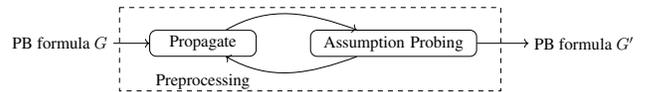

The preprocessing phase of {\pbcount} performs assumption probing and unit propagation~\cite{BJK21}. {\pbcount} repeatedly performs unit propagation and assumption probing until no change is detected, as shown in Algorithm~\ref{alg:preprocess}. 

\begin{algorithm}
\caption{Preprocess($G$)} \label{alg:preprocess}

\textbf{Input:} $G$ - PB formula

\textbf{Output: } $G'$ - preprocessed PB formula

\begin{algorithmic}[1]
\STATE mapping $\gets$ []; $G' \gets G$
\REPEAT
\FORALL{single variable constraint $C \in G'$}
    \STATE mapping $\gets$ mapping $\cup$ $\mathsf{InferDecision}(C)$
\ENDFOR
\STATE $G' \gets \mathsf{propagate}$($G'$, mapping)
\FORALL{variable $x \in G'$}
    \STATE mapping $\gets$ mapping $\cup$ $\mathsf{AssumProbe}(G',x)$
\ENDFOR
\STATE $G' \gets \mathsf{propagate}$($G'$, mapping)
\UNTIL{$G'$ does not change}
\RETURN $G'$

\end{algorithmic}
\end{algorithm}

\paragraph{Sign Manipulation}

Let $C$ be the PB constraint $-3x_1 - 4x_2 \leq -3$. One can multiply both sides of the constraint by $-1$ to form $3x_1 + 4x_2 \geq 3$. In addition, one would be able to switch the sign of the coefficient of $x_2$ as follows.
\begin{align*}
    &3x_1 + 4x_2 \geq 3 \\
    &3x_1 + 4(1-\bar{x}_2) \geq 3 \\
    &3x_1 - 4\bar{x}_2 \geq -1
\end{align*}
In general, one is able to manipulate the sign of any term coefficient as shown in the example above. We use the above technique to optimize PB constraint compilation approaches which we discuss in later sections.

\paragraph{Propagation}

Propagation in the Pseudo-Boolean context refers to the simplification of the PB constraints if decisions on some PB variables can be inferred. In particular, one might be able to infer decisions on PB variable $x_i$ from PB constraint $C_j$ when the constraint is of either 1) $a_i x_i \geq k$ or 2) $a_i x_i = k$ forms. We defer the details of the $\mathsf{InferDecision}$ algorithm to the Appendix. 

\begin{algorithm}
\caption{AssumProbe($G$, $x_i$)} \label{alg:assum-probe}

\textbf{Input:} $G$ - PB formula, $x_i$ - assumption variable

\textbf{Output: } mapping of variable values

\begin{algorithmic}[1]
\STATE temp, mapping $\gets []$ 
\FORALL{constraint $C \in G[x_i \mapsto 1]$}
    \STATE temp $\gets$ temp $\cup$ $\mathsf{InferDecision(C)}$
\ENDFOR

\FORALL{constraint $C \in G[x_i \mapsto 0]$}
    \STATE temp $\gets$ temp $\cup$ $\mathsf{InferDecision(C)}$
\ENDFOR

\FORALL{variable $x_j$, where $j \not = i$}
    \IF{exactly one literal of $x_j$ in temp}
        \STATE mapping $\gets$ mapping $\cup$ temp[$x_j$]
    \ENDIF
\ENDFOR
\RETURN mapping

\end{algorithmic}
\end{algorithm}

\paragraph{Assumption Probing}
Assumption probing can be viewed as a weaker form of failed literal probing~\cite{BJK21} as well as single step look ahead propagation process. For an arbitrary variable $x_i \in G$, where $G$ is the PB formula, assumption probing involves performing propagation and decision inference independently for when $x_i = 0$ and $x_i = 1$. If another variable $x_j$ is inferred to have the same value assignment $\tau[x_j]$ in both cases, then it can be inferred that $x_j$ should be set to $\tau[x_j]$ in all satisfying assignments of $G$. Algorithm~\ref{alg:assum-probe} illustrates the process for a single variable $x_i$, and in the preprocessing stage, we perform assumption probing on all variables in $G$.

\subsection{Pseudo-Boolean Constraint Compilation} \label{subsec:pb-clause-compile}

In this work, we introduce two approaches, namely top-down and bottom-up, to compile each constraint of a PB formula into an ADD. We use $T, k,$ and $eq$ in place of PB constraint $C$ when describing the compilation algorithms. $T$ refers to the term list, which is a list of $a_i x_i$ terms of $C$. $k$ is the constraint constant and $eq$ indicates if $C$ is `=' constraint.

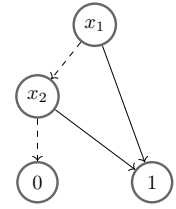
\begin{figure}[htb]
    \centering
    \begin{adjustbox}{width=0.25\columnwidth}
    \begin{tikzpicture}[
    roundnode/.style={circle, draw=black!60, very thick, minimum size=7mm},
    ]
    \node[roundnode](x1) at (0, 0){$x_1$};
    \node[roundnode](x2left) at (-1, -1.25){$x_2$};
    \node[roundnode](leaf0) at (-1, -2.75){$0$};
    \node[roundnode](leaf1) at (1, -2.75){$1$};

    \draw[dashed,->] (x1) -- (x2left);
    \draw[->] (x1) -- (leaf1);
    \draw[dashed,->] (x2left) -- (leaf0);
    \draw[->] (x2left) -- (leaf1);

    \end{tikzpicture}
    \end{adjustbox}
    \caption{An ADD $\psi_1$ representing $3x_1 + 4x_2 \geq 3$}
    \label{fig:add-expression-constraint}
\end{figure}

\paragraph{Bottom-up ADD Constraint Compilation}

In order to compile an ADD which represents a PB constraint of the following form $\sum_{i=1}^{n} a_i x_i [\geq, =, \leq] k$, we first start compiling the expression $\sum_{i=1}^{n} a_i x_i$ from literal and constant ADDs as shown by line~\ref{line:bot-up-compile:expression-add} of Algorithm~\ref{alg:bot-up-compile}. A constant ADD which represents integer $a_i$ is a single leaf node that has value $a_i$. A literal ADD comprises of an internal node, which represents variable $x$, and true and false leaf nodes, which represent the evaluated values of the literal if $x$ is set to true and false. With the literal and constant ADDs, we use {\apply} with $\times$ operator to form ADDs for each term $a_i x_i$. We use {\apply} with $+$ operator on term ADDs to form the ADD representing expression $\sum_{i=1}^{n} a_i x_i$. As an example, the ADD $\psi$ for the expression $3x_1 + 4x_2$ is shown in Figure~\ref{fig:add-expression}. To account for the inequality or equality, we look at the value of leaf nodes in expression ADD $\psi$ and determine if they satisfy the constraint (lines~\ref{line:bot-up-compile:constraint-sat-add-start} to~\ref{line:bot-up-compile:constraint-sat-add-end}). We replace the leaf nodes with $1$ node if the constraint is satisfied and $0$ node otherwise, the resultant ADD is illustrated in Figure~\ref{fig:add-expression-constraint}.

\begin{algorithm}
\caption{compileConstraintBottomUp($T, k, eq$)} \label{alg:bot-up-compile}

\textbf{Input:} $T$ - term list, $k$ - constraint value, 

$eq$ - indicator if constraint is `=' type

\textbf{Output: } $\psi$ - constraint ADD
\begin{algorithmic}[1]
    \STATE $\psi \gets \mathsf{constantADD}(0)$
    \FOR {term $t$ in $T$}
        \STATE $\psi \mathrel{+}= \mathsf{constantADD}$(t.coeff) $\mathrel{\times} \mathsf{literalADD}$(t.literal) \label{line:bot-up-compile:expression-add}
    \ENDFOR
    \FOR {node $n$ in $\mathsf{LeafNode}(\psi)$} \label{line:bot-up-compile:constraint-sat-add-start}
        \IF{$eq$ is \textit{true} \& $n$.value $= k$}
            \STATE $n$.value $\gets$ 1
        \ELSIF{$eq$ is \textit{false} \& $n$.value $\geq k$}
            \STATE $n$.value $\gets$ 1
        \ELSE
            \STATE $n$.value $\gets$ 0
            \label{line:bot-up-compile:constraint-sat-add-end}
        \ENDIF
    \ENDFOR
    \RETURN $\psi$
\end{algorithmic}
\end{algorithm}

\paragraph{Top-down ADD Constraint Compilation} 
In contrast to the bottom-up ADD compilation approach, the top-down ADD compilation for a given PB constraint involves the if-then-else ({\ite}) operation for decision diagrams. We only consider PB constraints that involve $=$ or $\geq$ as mentioned previously. 
The top-down compilation algorithm (Algorithm~\ref{alg:top-down-add-main}) makes use of recursive calls of Algorithm~\ref{alg:top-down-add-recur} to construct an ADD that represents a given PB clause. In particular, Algorithms~\ref{alg:top-down-add-main} and~\ref{alg:top-down-add-recur} work by iterating through the terms of the PB constraint using \textit{idx}. The algorithms build the sub-ADDs when the literal at position \textit{idx} evaluates to \textit{true} for the if-then case and otherwise for the else case of the {\ite} operation while updating the constraint constant $k$ (lines~\ref{line:top-down-add-main:lo-recur-call}-\ref{line:top-down-add-main:hi-recur-call} of Algorithm~\ref{alg:top-down-add-main} and lines~\ref{line:top-down-add-recur:lo-recur-call}-\ref{line:top-down-add-recur:hi-recur-call} of Algorithm~\ref{alg:top-down-add-recur}). Notice that the top-down compilation approach allows for early termination when the current $k$ value is negative for $\geq k$ case. However, early termination is possible only if all unprocessed coefficients are positive, implying that $k$ in subsequent recursive calls cannot increase. One way would be to sort the term list $T$ in ascending order of term coefficients, processing terms with negative coefficients before positive coefficients.

\begin{algorithm}
\caption{compileConstraintTopDown($T, k, eq$)} \label{alg:top-down-add-main}
\textbf{Assumption:} $T$ is in ascending order of term coefficients or all coefficients are non-negative

\textbf{Input:} $T$ - term list, $k$ - constraint value, 

$eq$ - indicator if constraint is `=' type

\textbf{Output: } $\psi$ - constraint ADD
\begin{algorithmic}[1]
    \STATE $\psi \gets \mathsf{literalADD}$($T$[0].literal)
    \STATE $\psi_{lo} \gets \mathsf{compileTDRecur}$($T, k, eq, 1$) \label{line:top-down-add-main:lo-recur-call}
    \STATE $\psi_{hi} \gets \mathsf{compileTDRecur}$($T, k - T$[0].coeff$, eq, 1$) \label{line:top-down-add-main:hi-recur-call}
    \STATE $\psi.\mathsf{ITE}$($\psi_{hi}, \psi_{lo}$)
    \RETURN $\psi$
\end{algorithmic}
\end{algorithm}

\begin{algorithm}
\caption{compileTDRecur($T, k, eq, idx$)} \label{alg:top-down-add-recur}
\textbf{Input:} $T$ - term list, $k$ - current constraint value, 

$eq$-input constraint equality, $idx$-index of current term in $T$

\textbf{Output: } $\psi$ - constraint ADD from $idx$ to end of $T$

\begin{algorithmic}[1]
    \IF{$T$[$idx$].coeff $\geq$ 0}
        \STATE $\mathsf{isPos} \gets true$
    \ENDIF
    \IF{$eq$ \& $\mathsf{isPos}$ \& $k < 0$}
        \RETURN $\mathsf{constantADD}(0)$
    \ELSIF{!$eq$ \& $\mathsf{isPos}$ \& $k \leq 0$}
        \RETURN $\mathsf{constantADD}(1)$
    \ELSIF{$idx < T$.length}
        \STATE $\psi \gets \mathsf{literalADD}$($T$[$idx$].literal)
        \STATE $\psi_{lo} \gets \mathsf{compileTDRecur}$($T, k, eq, idx + 1$) \label{line:top-down-add-recur:lo-recur-call}
        \STATE $\psi_{hi} \gets \mathsf{compileTDRecur}$($T, \newline 
        \hspace*{5em} k - T$[$idx$].coeff$, eq, idx + 1$) \label{line:top-down-add-recur:hi-recur-call}
        \RETURN $\psi.\mathsf{ITE}$($\psi_{hi}, \psi_{lo}$)
    \ELSE
        \IF{$eq$ \& $k = 0$}
            \RETURN $\mathsf{constantADD}(1)$
        \ELSE
            \RETURN $\mathsf{constantADD}(0)$
        \ENDIF
    \ENDIF
\end{algorithmic}
\end{algorithm}

\paragraph{Optimizations for Bottom-up Compilation} In the bottom-up compilation approach, an ADD is built from the individual literal and constant ADDs to represent the expression, before subsequently having leaf node values converted to $1$ and $0$ depending on if the PB constraint is satisfied. In the process, an ADD could be exponential in size with respect to the number of variables processed. In order to minimize the intermediate ADD during the compilation process, we introduce an optimization for bottom-up compilation. The key idea is to increase the number of shared sub-components of the intermediate ADD, and this amounts to processing the PB constraint terms in a manner that results in fewer distinct subset sums of term coefficients as every distinct subset sum requires a separate leaf node. To this end, we optimize the compilation process by sorting the terms according to the absolute values of their coefficients in ascending order. Subsequently, we manipulate the coefficients, using $x = (1-\bar{x})$, of the terms such that alternate terms have coefficients of different signs. We defer the pseudo code to the Appendix.

\paragraph{Optimizations for Top-down Compilation}
Similarly, we also introduce optimizations for the top-down compilation approach. Recall that one would only be able to perform early termination for PB constraints of the form $\sum a_i x_i \geq k$ after all negative coefficient terms have been processed. To this end, we manipulate all coefficients to be positive and adjust $k$ accordingly so that early termination is possible. Furthermore, we sort the terms in descending value of the term coefficients as larger coefficients are more likely to satisfy the constraint. We defer the pseudo code to the Appendix.

\begin{algorithm}
\caption{compileConstraintDynamic($T, k, eq$)} \label{alg:dynamic-compilation}
\textbf{Input:} $T$ - term list, $k$ - constraint value, $eq$-input constraint equality

\textbf{Output:} $\psi$ - constraint ADD

\textbf{Cond 1:} $T$.length $\leq$ 25 \textbf{and} $k < 25^{th}$ percentile of $T$.coeff

\textbf{Cond 2:} $k < 25^{th}$ percentile of $T$.coeff \textbf{and} unique coefficient rate $\geq$ 0.9 \textbf{and} unique adjacent difference rate $\geq$ 0.85

\begin{algorithmic}[1]
    \IF{cond 1 \textbf{or} cond 2}
        \STATE $\mathsf{bottomUp} \gets false$
    \ELSE
        \STATE $\mathsf{bottomUp} \gets true$
    \ENDIF
    \IF{$\mathsf{bottomUp}$}
        \RETURN $\mathsf{optimizeCompileBottomUp}(T, k, eq)$
    \ELSE
        \RETURN $\mathsf{optimizeCompileTopDown}(T, k, eq)$
    \ENDIF
\end{algorithmic}
\end{algorithm}

\paragraph{Dynamic Compilation} A PB formula can include more than one PB constraint. As we will show in a case study in the experiments section, the choice of compilation approach has a substantial impact on overall runtime. To this end, we introduce a dynamic heuristic (Algorithm~\ref{alg:dynamic-compilation}) to select the appropriate compilation approach and perform optimization of the compilation process as previously discussed. In Algorithm~\ref{alg:dynamic-compilation}, we choose top-down compilation if either condition 1 or 2 is met. Conditions 1 and 2 are designed to be in favor of the botttom-up compilation approach, we provide performance analysis in the experiments section.

\section{Experiments} \label{sec:experiments}

We performed extensive empirical evaluations to compare the runtime performance of {\pbcount} with state-of-the-art exact model counters. Our empirical evaluation focuses on benchmarks arising from three application domains: sensor placement, auctions, and multi-dimensional knapsack. 
Through our evaluations and analysis, we sought to answer the following research questions: 

\begin{description}
    \item[\textbf{RQ 1}] How does the runtime performance of  {\pbcount} compare to that of the state-of-the-art approaches? 
    \item[\textbf{RQ 2}] How does the  dynamic compilation approach impact the runtime performance of {\pbcount}?
\end{description}

\paragraph{Setup}

We performed our evaluations on machines with AMD EPYC 7713 processors. Each benchmark instance is provided with 1 core, 16GB memory, and a timeout of 3600 seconds. Since all the state-of-the-art exact model counters take CNF as input, we employed the CNF model counters with the help of PB to CNF conversion tool {\pblib}\footnote{We used the provided PBEncoder for conversion.}~\cite{PS15}. We evaluated {\pbcount} against state-of-the-art projected counters: {\dpmc}, {\dfour}\footnote{Binary from Model Counting Competition 2022} and {\gpmc}; {\dfour} and {\gpmc} are among the winners of the Projected counting track at Model Counting Competition 2022 and 2023. 

\paragraph{Benchmarks}

We generated 3473 benchmarks of the following application areas -- sensor placement, auctions, and multi-dimension knapsack. We detail the benchmark statistics (number of variables and constraints) in the Appendix.
\begin{itemize}
    \item The sensor placement benchmark setting (1473 instances after removal of 0 counts) is adapted from prior work on identifying code sets~\cite{LSM23}. Given a network graph, a maximum number of sensors allowed, count the number of ways to place sensors such that failures in the network are uniquely identifiable. 
    \item For the auction benchmark setting (1000 instances), we adapt the combinatorial auction setting~\cite{BN07} to a counting variant. There are $m$ participants and $n$ items, each of which can be shared by one or more participants. Given that each participant has a minimum utility threshold, we count the number of ways the $n$ items can be shared such that all participants achieve their minimum threshold. The utilities are additive and can be negative.
    \item For the multi-dimension knapsack benchmark setting~\cite{GL80} (1000 instances), there are $n$ items and constraints on $m$ different features or dimensions of the items in the form of the sum of each dimension should not exceed a given constant. Given such a setting, the goal would be the count the number of subsets of items that satisfies the constraints.
\end{itemize}

\subsection{RQ1: Runtime Comparison}

\begin{figure}[htb]
\centering
\includegraphics[width=0.9\linewidth]{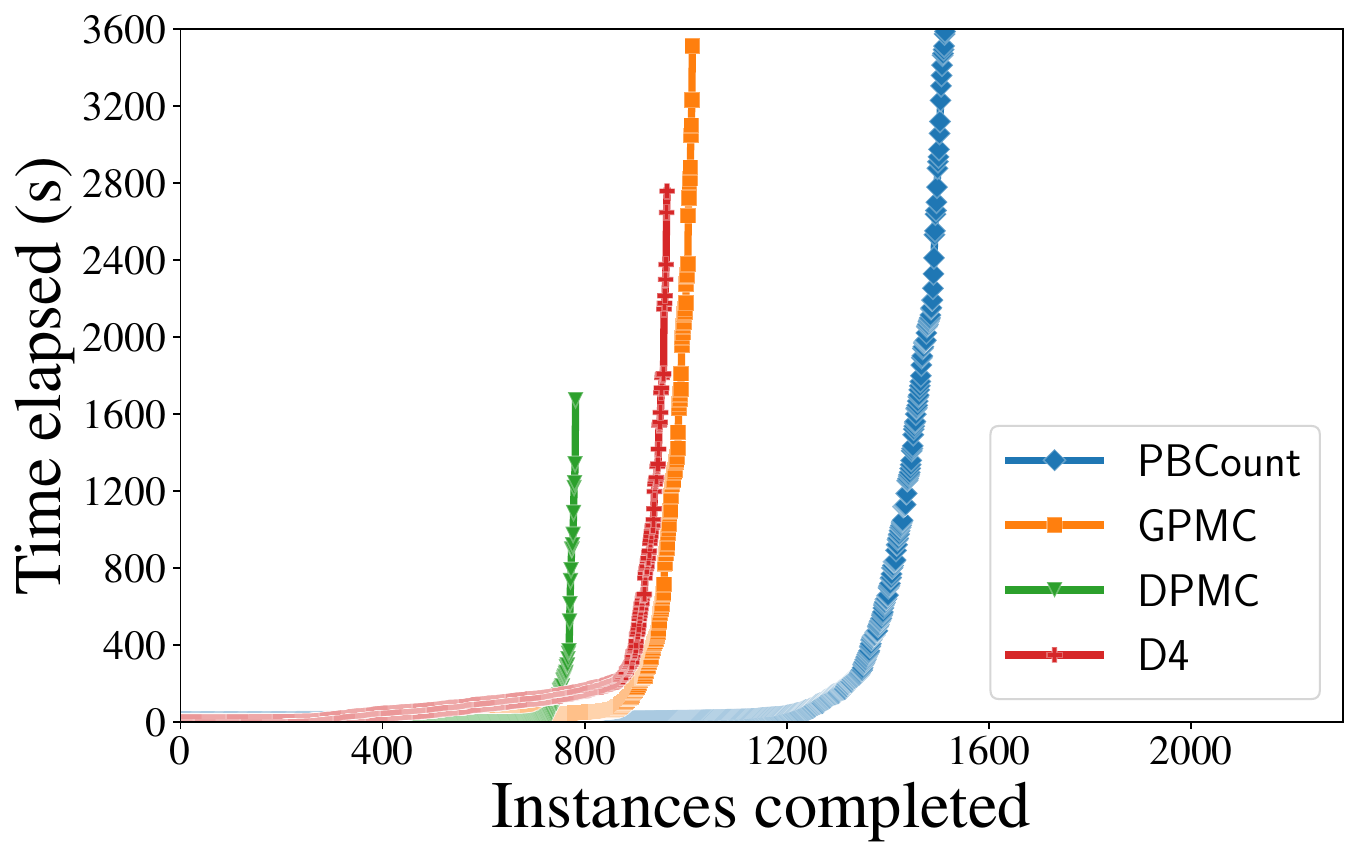}
\caption{Cactus plot of number of benchmark instances completed by different counters. A point ($x,y$) on each line plot indicates the corresponding counter completes $x$ number of benchmarks after $y$ seconds has elapsed.}
\label{fig:overall-cactus-plot}
\end{figure}

We show the cactus plot of the number of instances completed by each counter out of the 3473 benchmarks in Figure~\ref{fig:overall-cactus-plot}. The exact number of instances completed by each counter for each benchmark set is shown in Table~\ref{tab:benchmark-overall}. Additionally, we provide individual cactus plots for each set of benchmarks in the Appendix.

\begin{table}[h!]
    \centering
    \begin{small}
    \begin{NiceTabular}{l|r|r|r|r}
    \toprule
    Benchmarks                      & {\dpmc}            & {\dfour}         & {\gpmc}           & {\pbcount}        \\
    \midrule
    Sensor placement                & 625                & 566              & 575               & \textbf{638}      \\
    $\mathcal{M}$-dim knapsack      & 81                 & 281              & 279               & \textbf{503}      \\
    Auction                         & 76                 & 116              & 159               & \textbf{372}      \\
    \midrule
    \midrule
    Total                           & 782                & 963              & 1013              & \textbf{1513}     \\
    \bottomrule
    \end{NiceTabular}
    \end{small}
    \caption{Number of benchmark instances completed by each counter in 3600s, higher is better.}
    \label{tab:benchmark-overall}
\end{table}

In sensor placement benchmarks, {\pbcount} count completed 638 instances, narrowly ahead of {\dpmc} (625 instances), and more than {\dfour} (566 instances) and {\gpmc} (575 instances). In multi-dimension knapsack ($\mathcal{M}$-dim knapsack) and auction benchmarks, {\pbcount} significantly outperforms the competing counters. {\pbcount} completed 503 $\mathcal{M}$-dim knapsack instances, around 1.8$\times$ that of {\gpmc} (279 instances) and {\dfour} (281 instances), and 6.2$\times$ that of {\dpmc} (81 instances). In auction benchmarks, {\pbcount} completed 372 instances, around 2.3$\times$ that of {\gpmc} (159 instances), 3.2$\times$ of {\dfour} (116 instances), and 4.9$\times$ of {\dpmc} (76 instances). Overall, {\pbcount} completed 1513 instances out of 3473 total instances, around 1.5$\times$ that of {\gpmc}, 1.6$\times$ of {\dfour}, and 1.9$\times$ that of {\dpmc}. Note that {\pbcount} achieved superior performance with minimal preprocessing over {\gpmc}, which has advanced preprocessing capabilities. Our results demonstrate the significant performance advantages of counting natively for PB formulas and provide an affirmative answer to \textbf{RQ1}.

\subsection{RQ2: Analysis of Compilation Approaches}

\begin{figure}[htb]
    \centering
    \includegraphics[width=0.75\columnwidth]{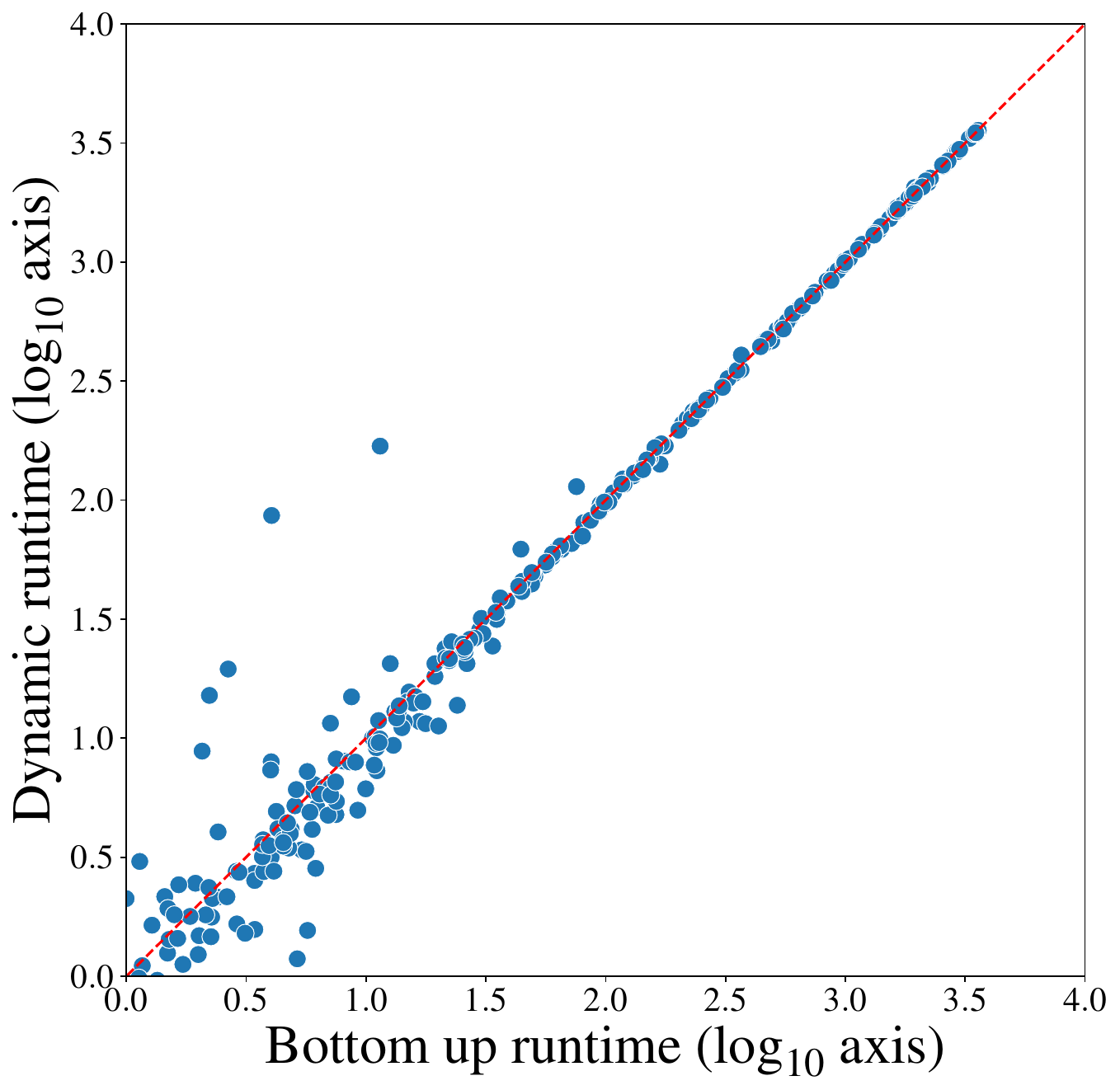}
    \caption{Dynamic vs bottom-up runtime ($\log_{10}$) for auction benchmarks. Points beneath red diagonal line indicate dynamic compilation is faster (257 points), points above otherwise (115 points).}
    \label{fig:bu-vs-dynamic-runtime-mdim}
\end{figure}

We now focus on the analysis of different compilation approaches: top-down (Algorithm~\ref{alg:top-down-add-main}), bottom-up (Algorithm~\ref{alg:bot-up-compile}), and dynamic (Algorithm~\ref{alg:dynamic-compilation}). The results in Table~\ref{tab:benchmark-compilation-choice} show that for the benchmarks, bottom-up PB constraint compilation outperforms top-down approach significantly in auction and multi-dimension knapsack and to a lesser degree sensor placement. In addition, the evaluation result also highlights that our dynamic compilation heuristic and constraint term optimization closely match the bottom-up approach, with the exception of completing 3 fewer instances in auction benchmarks. However, in the 372 auction instances completed by both bottom-up and dynamic approaches, the dynamic approach with term coefficient optimization completes the counting task faster for 257 instances. We show the scatter plot comparison in Figure~\ref{fig:bu-vs-dynamic-runtime-mdim}.

\begin{table}[htb]
\centering
\begin{small}
    \begin{NiceTabular}{l|r|r|r}
    \toprule
    Benchmarks          & Top-down             & Bottom-up            & Dynamic              \\
    \midrule
    Sensor placement    & 580                  & 638                  & 638                  \\
    $\mathcal{M}$-dim  knapsack      & 109                  & 503                  & 503                  \\
    Auction             & 158                  & 375                  & 372                  \\
    \bottomrule
    \end{NiceTabular}
\end{small}
\caption{Number of benchmarks completed by {\pbcount} when employing different compilation strategies, higher number indicates better performance.}
\label{tab:benchmark-compilation-choice}
\end{table}

\paragraph{Compilation Approach Performance Case Study}
We provide an example to highlight the performance impact of the choice of compilation approach. The example involves the following PB formula in Equation~\ref{eq:casestudy-eqn} with a single constraint that has unique term coefficients:
\begin{equation} \label{eq:casestudy-eqn}
    \sum_{i=0}^{12} 2^{i} x_{i+1} + \sum_{i=1}^{10} 3^{i} x_{i+13} + \sum_{i=1}^{7} 7^{i} x_{i+23} \geq k
\end{equation}
We vary the value of $k$ in the above PB constraint from $10^1$ to $10^5$ and compare the runtime between top-down and bottom-up compilation approaches in Table~\ref{tab:casestudy-vary-k}. Note that bottom-up compilation takes around the same time irrespective of $k$ as there is no early termination. On the other hand for top-down compilation, the PB constraint is easily satisfied when $k$ is small and thus allows for early termination, leading to significant time savings compared to when $k$ is large. Notice that when top-down compilation is unable to terminate early, it is much slower than bottom-up compilation even when all term coefficients are unique.

\begin{table}[htb]
\centering
\begin{small}
    \begin{NiceTabular}{l|r|r|r|r|r}
    \toprule
    \multirow{2}{*}{Approach} & \multicolumn{5}{c}{$k$ value} \\
    \cmidrule(l{0em}){2-6}
    & $10^1$ & $10^2$ & $10^3$ & $10^4$ & $10^5$ \\
    \midrule
    Top-down            & 0.005 & 0.009 & 0.228 & 8.586 & 46.071 \\
    Bottom-up           & 6.927 & 7.202 & 7.198 & 7.434 & 6.732 \\
    \bottomrule
    \end{NiceTabular}
\end{small}
\caption{Runtime (seconds) to complete model counting for formula in Equation~\ref{eq:casestudy-eqn}. Lower is better}
\label{tab:casestudy-vary-k}
\end{table}

\begin{table}[htb]
\centering
\begin{small}
    \begin{NiceTabular}{l|r|r|r|r|r}
    \toprule
    \multirow{2}{*}{Approach} & \multicolumn{5}{c}{$k$ value} \\
    \cmidrule(l{0em}){2-6}
    & $10^1$ & $10^2$ & $10^3$ & $10^4$ & $10^5$ \\
    \midrule
    Top-down            & 3.325 & 61.753 & 60.530 & 60.881 & 64.097 \\
    Bottom-up           & 0.005 & 0.004 & 0.004 & 0.004 & 0.004 \\
    \bottomrule
    \end{NiceTabular}
\end{small}
\caption{Runtime (seconds) to complete model counting for formula in Equation~\ref{eq:casestudy-eqn} with all coefficients set to 1.}
\label{tab:casestudy-vary-k-same-coeff}
\end{table}

As mentioned previously, bottom-up compilation benefits from having large numbers of same term coefficients or collisions in subset sums of coefficients. To this end, we changed all term coefficients of the PB constraint in equation~\ref{eq:casestudy-eqn} to $1$ and compared runtimes in Table~\ref{tab:casestudy-vary-k-same-coeff}. We observed around three orders of magnitude reduction in the runtime of the bottom-up compilation approach. In contrast, the top-down approach terminates early only in $k=10^1$ case and requires full enumeration in other cases. In the absence of early termination, top-down compilation approach is much slower than bottom-up compilation approach, and this is reflected in our dynamic compilation heuristic.

\section{Conclusion} \label{sec:conclusion}

In this work, we introduce the first exact PB model counter, {\pbcount}. {\pbcount} directly compiles PB formulas into ADDs, enabling us to reuse the ADD counting framework in {\addmc}. In the design of {\pbcount}, we introduce both top-down and bottom-up PB constraint compilation techniques and highlight the performance differences between them. While we introduced dynamic compilation heuristics to determine the per constraint compilation method and preliminary preprocessing techniques for PB formulas, it would be of interest to develop more advanced heuristics and preprocessing techniques in future works. A strong motivation is {\pbcount}'s performance lead over existing CNF model counters. We hope this work will gather more interest in PB formulas and PB model counting.

\section*{Acknowledgments}
The authors thank Anna L.D. Latour for helping during benchmark generation. The authors thank Arijit Shaw and Jiong Yang for constructive discussions. The authors thank reviewers for providing feedback. This work was supported in part by the Grab-NUS AI Lab, a joint collaboration between GrabTaxi Holdings Pte. Ltd. and National University of Singapore, and the Industrial Postgraduate Program (Grant: S18-1198-IPP-II), funded by the Economic Development Board of Singapore. This work was supported in part by National Research Foundation Singapore under its NRF Fellowship Programme [NRF-NRFFAI1-2019-0004], Ministry of Education Singapore Tier 2 grant [MOE-T2EP20121-0011], and Ministry of Education Singapore Tier 1 Grant [R-252-000-B59-114]. The computational work for this article was performed on resources of the National Supercomputing Centre, Singapore.

\bibliography{references}
\appendix
\clearpage
\onecolumn

\section*{Appendix}

\subsection*{Algorithm Details}

\begin{algorithm}
\caption{InferDecision($C$)} 
\label{app:alg:prop-infer-decision}

\textbf{Input:} $C$ - PB constraint with single term $a_i l_i$ where $l_i$ is a literal of variable $x_i$

\textbf{Output: } assignment mapping of $x_i$ if inferred

\begin{algorithmic}[1]
\STATE mapping $\gets []$ 
\IF{$C$ is equality}
    \IF{$a_i = k$ and $k \not = 0$}
        \STATE mapping $\gets \mathsf{mapLitTrue(C)}$
    \ELSIF{$k = 0$}
        \STATE mapping $\gets \mathsf{mapLitFalse(C)}$
    \ENDIF
\ELSE
    \IF{($0 < k \leq a_i$ and $\mathsf{isPosLit}(l_i)$) \\ or ($a_i \leq k < 0$ and $\mathsf{isNegLit}(l_i)$)}
        \STATE mapping $\gets \mathsf{mapLitTrue(C)}$
    \ENDIF
\ENDIF
\RETURN mapping

\end{algorithmic}
\end{algorithm}

As mentioned in the main paper, it could be possible to infer decision in variables for PB constraints that comprises of only one literal, or one term. We present the algorithm for inferring decisions as Algorithm~\ref{app:alg:prop-infer-decision}, $\mathsf{InferDecision}$. Given a PB constraint $C$, there are two cases -- when $C$ is an equality and when $C$ is of $\geq$ form. When $C$ is an equality, $C$ is satisfied either when its constraint constant $k = 0$ (we infer literal $l_i$ should be \textit{false}) or when $a_i = k$ (we infer literal $l_i$ should be \textit{true}). When $C$ is an inequality of $\geq$ form, we can infer that literal $l_i$ should be \textit{true} in two scenarios. The first is when $l_i$ is not a negated literal and that $0< k \leq a_i$. The second is when $l_i$ is a negated literal and $a_i \leq k < 0$. In other cases, we are unable to infer a decision for variable $x_i$.

\begin{algorithm}
\caption{optimizeCompileBottomUp($T, k, eq$)} 
\label{app:alg:optimize-bot-up}
\textbf{Input:} $T$ - term list, $k$ - constraint value, $eq$ - input constraint equality

\textbf{Output: } $\psi$ - constraint ADD
\begin{algorithmic}[1]
    \STATE $T \gets \mathsf{sortAscendingAbsoluteCoeff}(T)$
    \STATE $T', k' \gets \mathsf{makeEveryAltCoeffPos}(T,k)$
    \RETURN $\mathsf{compileConstraintBottomUp}(T', k', eq)$
\end{algorithmic}
\end{algorithm}

In the main paper, Section~\ref{sec:approach}, we described optimizations for the PB constraint compilation process. In particular, we mentioned that for bottom-up compilation approach, it would be desirable if the intermediate ADD is small when compiling the terms component of the given PB constraint. To that end, we show our optimizations for bottom-up constraint ADD compilation in Algorithm~\ref{app:alg:optimize-bot-up}. We sort the terms in the PB constraint in ascending term coefficient magnitude, and subsequently manipulate the constraint such that adjacent term coefficients have different signs. The idea is to improve the collision rate of subset sums of coefficients so that there are fewer leaf nodes required in the intermediate ADD.

\begin{algorithm}
\caption{optimizeCompileTopDown($T, k, eq$)} 
\label{app:alg:optimize-top-down}
\textbf{Input:} $T$ - term list, $k$ - constraint value, $eq$ - input constraint equality

\textbf{Output: } $\psi$ - constraint ADD
\begin{algorithmic}[1]
    \STATE $T', k' \gets \mathsf{makeAllCoeffPos}(T,k)$
    \STATE $T' \gets \mathsf{sortDescendingCoeff}(T)$
    \RETURN $\mathsf{compileConstraintTopDown}(T', k', eq)$
\end{algorithmic}
\end{algorithm}

Similarly, we introduce optimizations for top-down constraint ADD compilation process in Algorithm~\ref{app:alg:optimize-top-down}. The main optimization idea for top-down compilation approach is to process the terms in such a manner that allows for early termination. As mentioned in the main paper, early termination can happen when PB constraint is of $\geq$ form. In particular, early termination can only take place after all terms with negative coefficients have been processed. To this end, we manipulate all term coefficients to be positive so that early termination can take place at all times. In addition, processing larger coefficients will satisfy the $\geq$ inequality sooner. As such, our optimizations included processing the terms in decending order of coefficients.

\subsection*{Benchmark Statistics}

In this work, we generated benchmarks for various applications -- sensor placement, multidimension knapsack, and combinatorial auction applications. We detail the number of variables, clauses, and constraints of the PB formula and converted CNF formula, in the rest of this section.

\begin{table}[H]
    \centering
    \begin{small}
    \begin{NiceTabular}{l|r|r|r|r}
    \toprule
    Statistics          & \# PB variable        & \# PB constraint    & \# CNF variable      & \# CNF clause    \\
    \midrule
    Min                 & 5                     & 1                   & 5                    & 1                \\
    25\%                & 60                    & 4                   & 15830                & 28456            \\
    50\%                & 91                    & 9                   & 41900.5              & 77265.5          \\
    75\%                & 131                   & 14                  & 80270                & 147022.75        \\
    Max                 & 252                   & 20                  & 234777               & 474900\\
    \bottomrule
    \end{NiceTabular}
    \end{small}
    \caption{Statistics of number variables, number of constraints and number of clauses for PB and CNF formula in the auction benchmarks}
    \label{app:tab:auction-stats}
\end{table}

Table~\ref{app:tab:auction-stats} shows the benchmark statistics for the auction setting. We show the minimum number, $25^{th}$ percentile, median, $75^{th}$ percentile, and max value for the number of variables, constraints, and clauses for both PB formula and converted CNF formula. The `\# PB variable' and `\# PB constraint' columns show statistics for the number of variables and constraints in the 1000 PB formulas (each PB formula is one benchmark), on median (`50\%' row) the PB benchmarks have 91 variables and 9 constraints. Similarly, the numbers for the converted CNF formulas are shown under `\# CNF variable' and `\# CNF clause' columns. The CNF version of the benchmarks have on median 41900.5 variables and 77265.5 clauses, which is 4 to 5 orders of magnitude larger than that of the PB version, this strongly supports our claim that PB formulas are polynomially more succinct than CNF formulas. The observation also holds for $\mathcal{M}$-dim knapsack benchmarks (multidimension knapsack) which we show the stats for in Table~\ref{app:tab:m-dim-knapsack-stats}.

\begin{table}[H]
    \centering
    \begin{small}
    \begin{NiceTabular}{l|r|r|r|r}
    \toprule
    Statistics          & \# PB variable        & \# PB constraint    & \# CNF variable      & \# CNF clause    \\
    \midrule
    Min                 & 5                     & 1                   & 28                   & 38               \\
    25\%                & 86.75                 & 6                   & 7866                 & 14871.25         \\
    50\%                & 164                   & 10                  & 25212                & 46340            \\
    75\%                & 234                   & 15                  & 50670.5              & 92525.25         \\
    Max                 & 300                   & 20                  & 171632               & 314821           \\
    \bottomrule
    \end{NiceTabular}
    \end{small}
    \caption{Statistics of number variables, number of constraints and number of clauses for PB and CNF formula in the $\mathcal{M}$-dim knapsack benchmarks}
    \label{app:tab:m-dim-knapsack-stats}
\end{table}

However, the difference is not as large in sensor placement benchmarks, because the coefficients and $k$ values are predominantly 1 and -1 except for the budget constraint which indicates the maximum number of sensors one could place. As such the sensor placement benchmarks are also more amenable to CNF encodings, unlike auction and knapsack benchmarks with different coefficients. We show the benchmark statistics for sensor placement setting in Table~\ref{app:tab:sensor-placement-stats}.

\begin{table}[H]
    \centering
    \begin{small}
    \begin{NiceTabular}{l|r|r|r|r}
    \toprule
    Statistics          & \# PB variable        & \# PB constraint    & \# CNF variable      & \# CNF clause    \\
    \midrule
    Min                 & 1                     & 2                   & 1                    & 1                \\
    25\%                & 25                    & 291                 & 158                  & 539.75           \\
    50\%                & 81.5                  & 1210                & 1154                 & 3745.5           \\
    75\%                & 190                   & 7056.5              & 3932                 & 12902.5          \\
    Max                 & 300                   & 44552               & 7699                 & 55514            \\
    \bottomrule
    \end{NiceTabular}
    \end{small}
    \caption{Statistics of number variables, number of constraints and number of clauses for PB and CNF formula in the sensor placement benchmarks}
    \label{app:tab:sensor-placement-stats}
\end{table}

\subsection*{Additional Results}

\begin{figure*}[htb]
\begin{subfigure}{0.33\textwidth}
    \includegraphics[width=\linewidth]{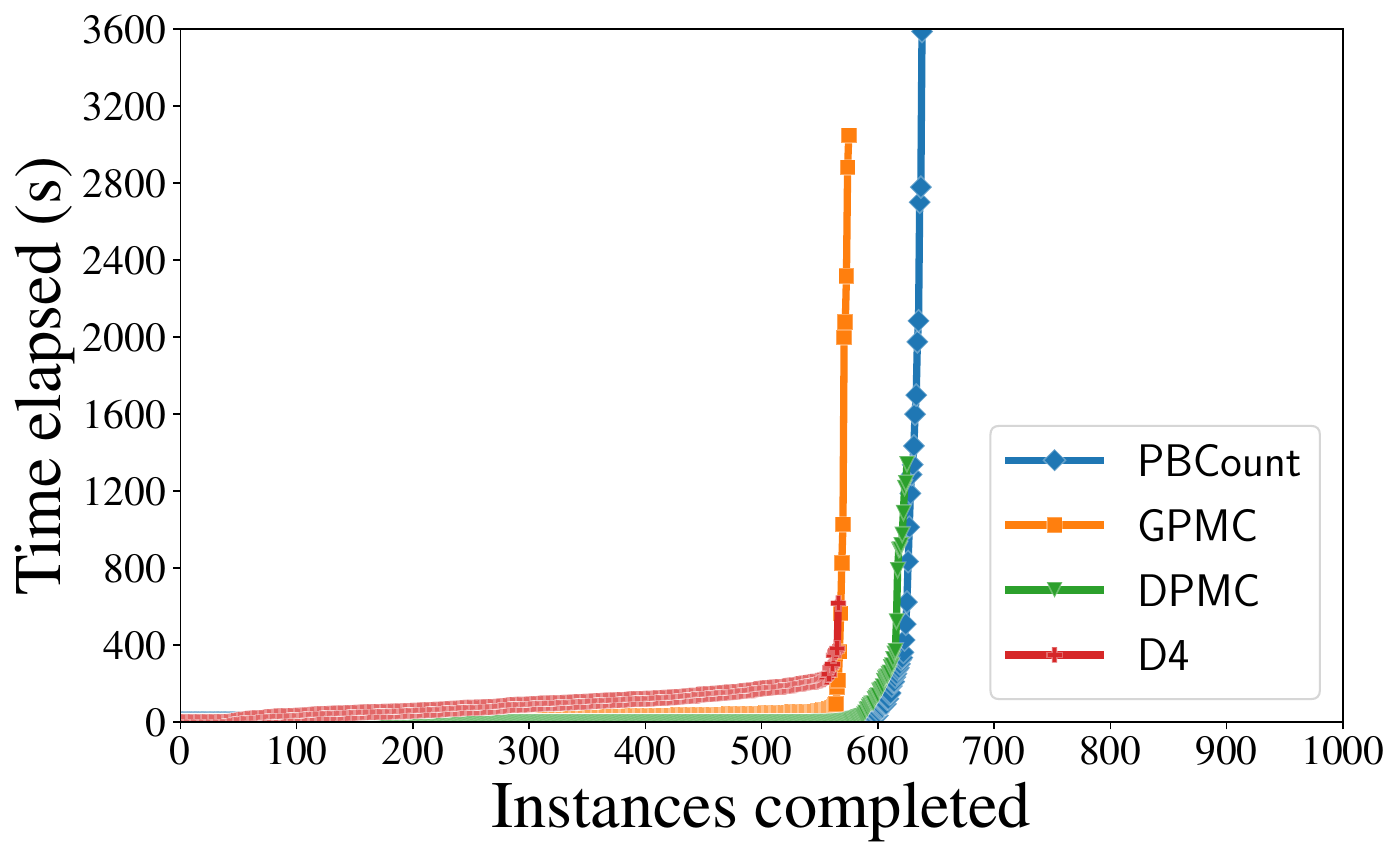}
    \caption{Sensor placement} \label{app:fig:sensor-placement}
\end{subfigure}%
\hfill
\begin{subfigure}{0.33\textwidth}
    \includegraphics[width=\linewidth]{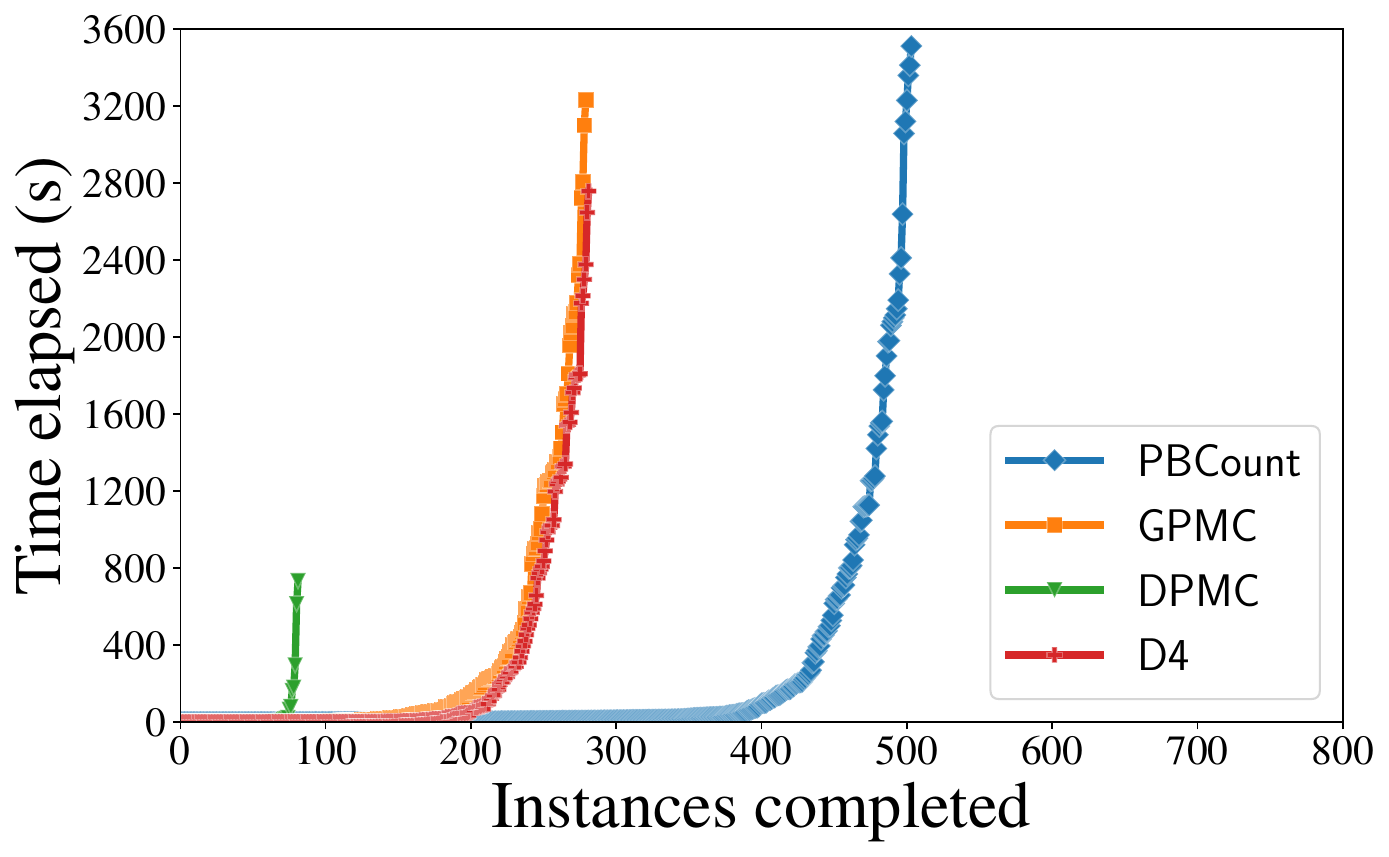}
    \caption{$\mathcal{M}$-dim Knapsack} \label{app:fig:m-dim-knapsack}
\end{subfigure}%
\hfill
\begin{subfigure}{0.33\textwidth}
    \includegraphics[width=\linewidth]{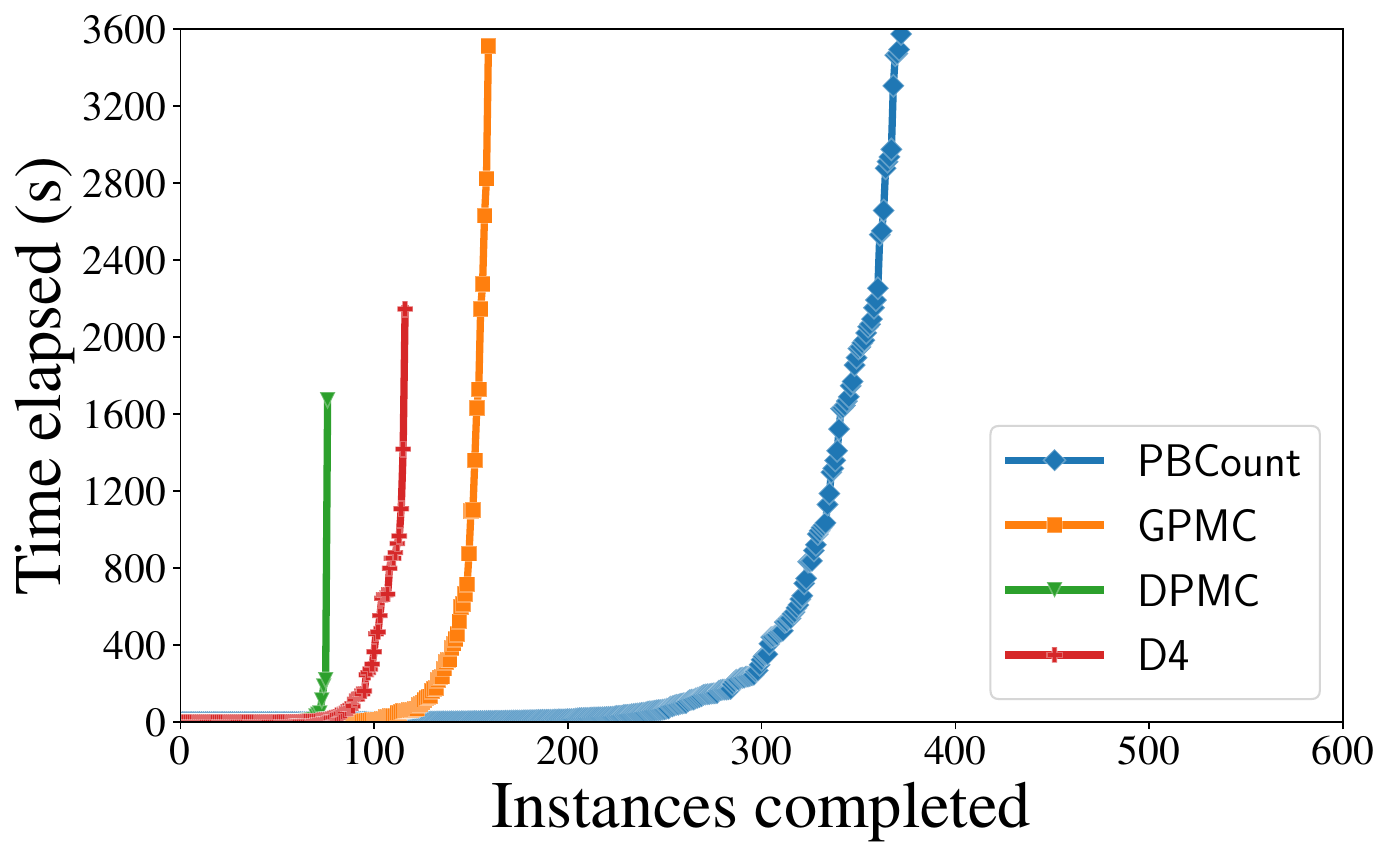}
    \caption{Auction} \label{app:fig:auction-cactus}
\end{subfigure}
\caption{Cactus plots of different benchmark sets. A point ($x,y$) on each line plot indicates the corresponding counter completes $x$ number of benchmarks after $y$ seconds has elapsed.}
\label{app:fig:individual-cactus-plot}
\end{figure*}

In the main paper, we provided a cactus plot showing the total number of benchmark instances completed by each counter. We provide the cactus plots for each set of benchmarks in Figure~\ref{app:fig:individual-cactus-plot}.

In the main paper, we showed the number of benchmarks completed by each counter. We show the number of unique benchmarks completed by each counter, along with the complete results in Table~\ref{app:tab:benchmark-overall-unique}.

\begin{table}[H]
    \centering
    \begin{small}
    \begin{NiceTabular}{l|r|r|r|r}
    \toprule
    Benchmarks                      & {\dpmc}            & {\dfour}         & {\gpmc}           & {\pbcount}        \\
    \midrule
    Sensor placement                & 625 (2)            & 566 (0)          & 575 (0)           & \textbf{638} (\textbf{11})     \\
    $\mathcal{M}$-dim knapsack      & 81 (0)             & 281 (13)         & 279 (2)           & \textbf{503} (\textbf{242})     \\
    Auction                         & 76 (0)             & 116 (0)          & 159 (12)          & \textbf{372} (\textbf{229})     \\
    \midrule
    \midrule
    Total                           & 782 (2)            & 963 (13)         & 1013 (14)         & \textbf{1513} (\textbf{482})    \\
    \bottomrule
    \end{NiceTabular}
    \end{small}
    \caption{Number of benchmark instances completed by each counter in 3600s, higher is better. Number in brackets indicates number of unique instances completed by the respective solver.}
    \label{app:tab:benchmark-overall-unique}
\end{table}

\begin{figure*}[htb]
    \begin{subfigure}{0.3\textwidth}
        \includegraphics[width=\linewidth]{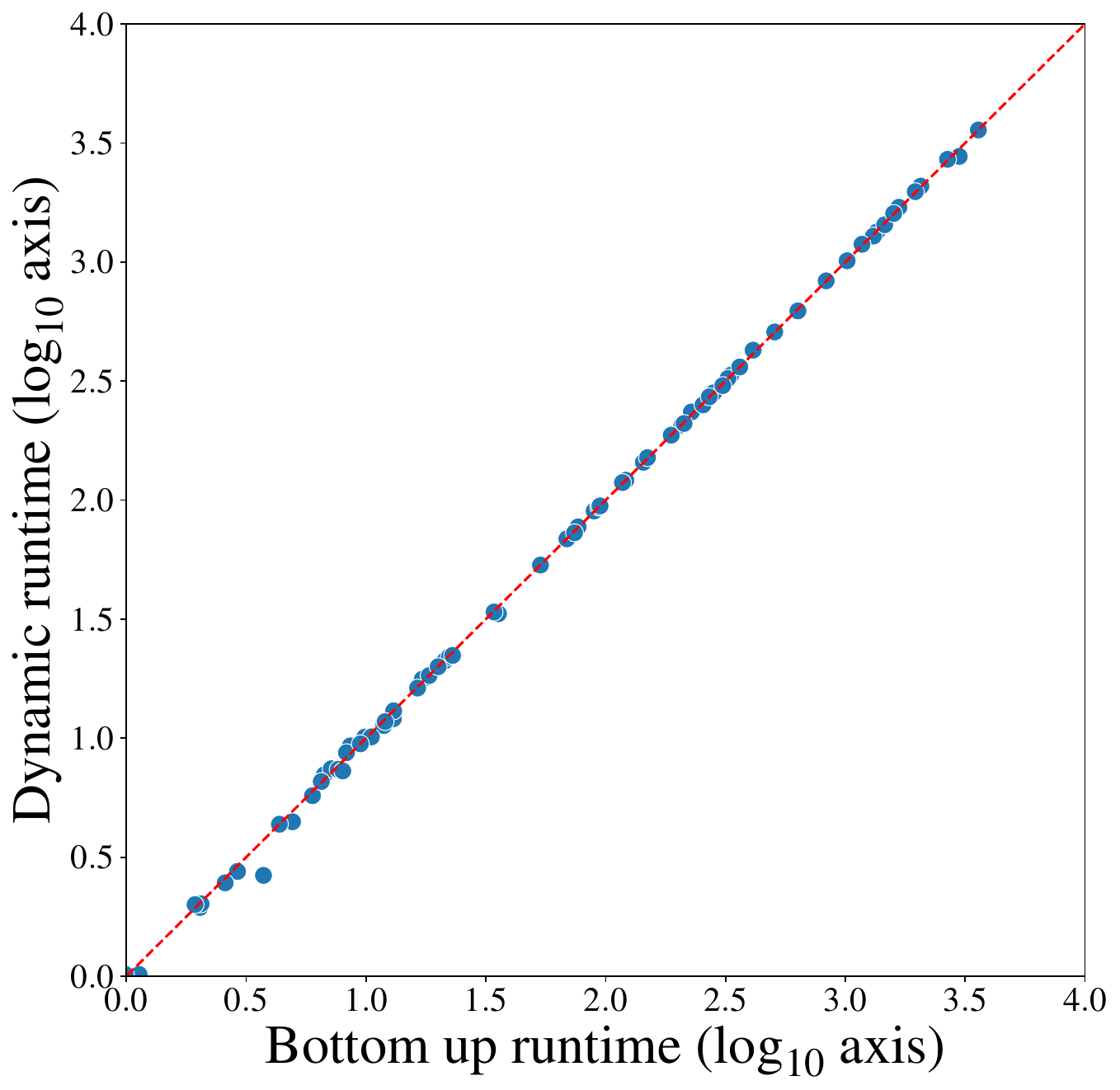}
        \caption{Sensor placement}
    \end{subfigure}%
    \hfill
    \begin{subfigure}{0.3\textwidth}
        \includegraphics[width=\linewidth]{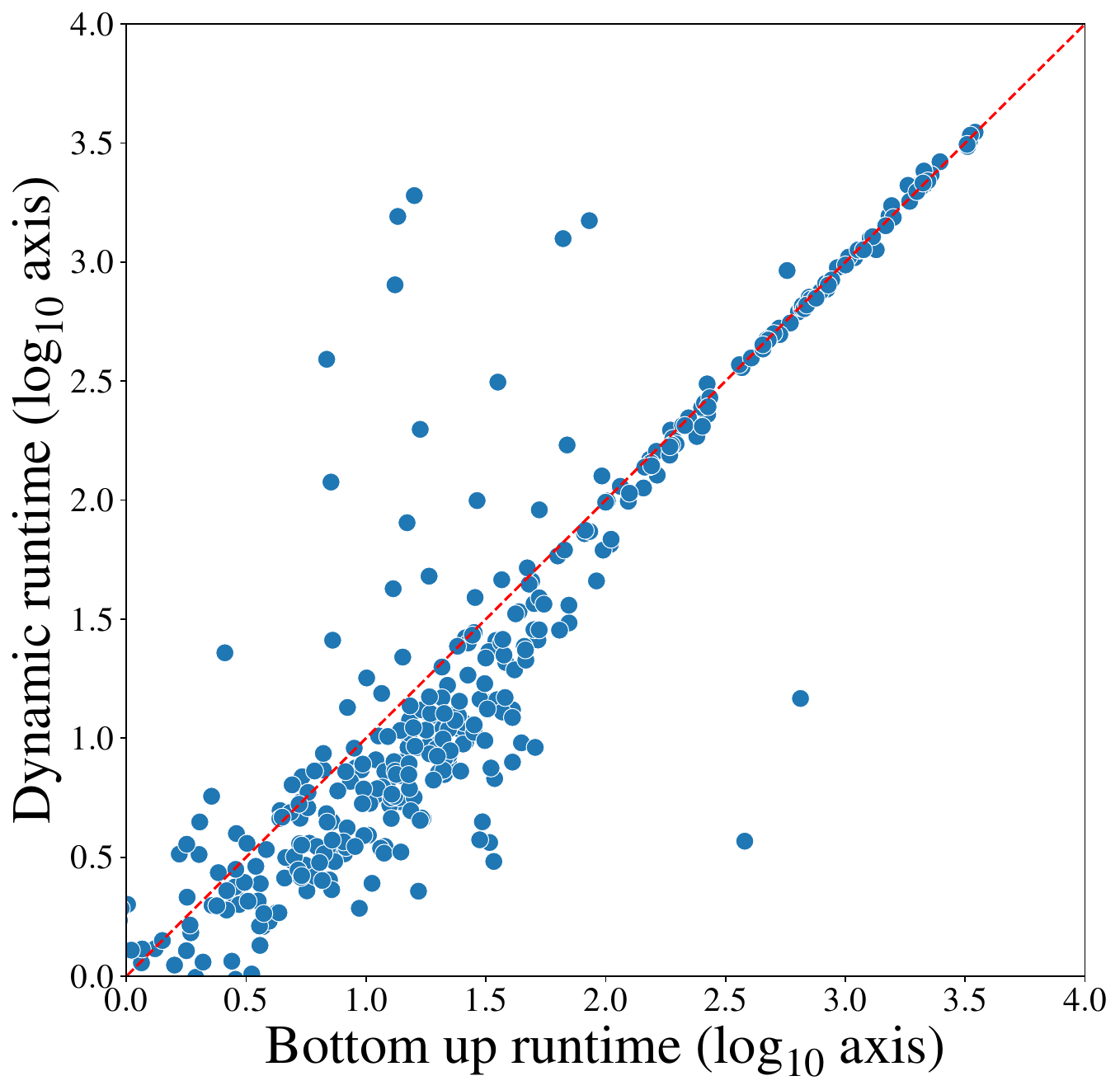}
        \caption{$\mathcal{M}$-dim Knapsack}
    \end{subfigure}%
    \hfill
    \begin{subfigure}{0.3\textwidth}
        \includegraphics[width=\linewidth]{diagrams/auction-botup-vs-dynamic-scatter-3600s.pdf}
        \caption{Auction}
    \end{subfigure}
    \caption{Scatter plots of runtimes of different benchmark sets between bottom-up and dynamic compilation approaches. Points beneath red diagonal line indicate dynamic compilation is faster, points above otherwise.}
    \label{app:fig:individual-scatter-plot-bu-dyn}
\end{figure*}

In the main paper, we showed the scatter plot of auction benchmark setting runtimes between bottom-up and dynamic compilation. We show the scatter plot of all three benchmark setting runtimes between the bottom-up and dynamic compilation approaches in Figure~\ref{app:fig:individual-scatter-plot-bu-dyn}.

In the main paper, we mentioned that {\pbcount} outperforms state-of-the-art propositional model counter {\gpmc} while using minimal preprocessing techniques relative to {\gpmc}'s advanced preprocessing techniques. We show the number of benchmarks completed by both {\pbcount} (dynamic compilation) and {\gpmc} with and without preprocessing in Table~\ref{app:tab:preprocessing-ablation} with 3600 seconds timeout. We were not able to completely turn off preprocessing and simplification in {\gpmc}, we only turned it off using the preprocessing parameter flags exposed to users by {\gpmc}.

\begin{table}[H]
\centering
\begin{small}
    \begin{NiceTabular}{l|r|r|r|r}
    \toprule
    Benchmarks                      & {\gpmc} (no-pp)     & {\gpmc}          & {\pbcount} (no-pp)            &  {\pbcount}              \\
    \midrule
    Sensor placement                & 573                 & 575              & 635              & \textbf{638}  \\
    $\mathcal{M}$-dim knapsack      & 268                 & 279              & \textbf{504}     & 503           \\
    Auction                         & 125                 & 159              & 371              & \textbf{372}  \\
    \midrule
    \midrule
    Total                           & 966                 & 1013             & 1510             & \textbf{1513} \\
    \bottomrule
    \end{NiceTabular}
\end{small}
\caption{Number of benchmarks completed by each counter with and without preprocessing (`no-pp' indicates no preprocessing) with 3600s timeout, higher is better.}
\label{app:tab:preprocessing-ablation}
\end{table}

Without preprocessing, {\pbcount} completed 3 fewer benchmark instances in total while {\gpmc} showed a significant performance drop by completing 47 fewer instances. We further highlight the performance impact of preprocessing to {\pbcount} and {\gpmc} by showing the runtime statistics in Table~\ref{app:tab:preprocessing-ablation-runtime}.

\begin{table}[H]
    \centering
    \begin{small}
    \begin{NiceTabular}{l|r|r|r|r}
    \toprule
    Runtime statistics  & {\gpmc} (no-pp)       & {\gpmc}             & {\pbcount} (no-pp)   &  {\pbcount}  \\
    \midrule
    Min                 & 0.00                  & 0.00                & 0.00                 & 0.00         \\
    25\%                & 11.23                 & 3.94                & 0.21                 & 0.19         \\
    50\%                & 51.27                 & 19.34               & 0.76                 & 0.66         \\
    75\%                & 110.43                & 42.10               & 16.25                & 14.92        \\
    Max                 & 3564.52               & 3513.57             & 3584.46              & 3587.59      \\
    \bottomrule
    \end{NiceTabular}
    \end{small}
    \caption{Runtime statistics of number benchmark instances completed by {\pbcount} and {\gpmc}, `(no-pp)' corresponds to settings where preprocessing of counter is disabled.}
    \label{app:tab:preprocessing-ablation-runtime}
\end{table}

Additionally, we conducted experiments to further analyze the performance of top-down and bottom-up compilation approaches. In particular, we modified the auction benchmarks in two ways (a) for each PB constraint we set the term coefficient to $\pm a$ where $a$ is randomly generated, and (b) for each PB constraint we set the term coefficient to one of three randomly generated value independently ($\pm a, \pm b$ or $\pm c$). We show the results in Table~\ref{app:tab:coeff-modification-auction-result}.  

\begin{table}[H]
\centering
\begin{small}
    \begin{NiceTabular}{l|r|r|r}
    \toprule
    Benchmarks                  & Top-down             & Bottom-up            & Dynamic              \\
    \midrule
    Same value (887)      & 236                  & 758                  & 757                  \\
    3 values (903)        & 258                  & 705                  & 705                  \\
    \bottomrule
    \end{NiceTabular}
\end{small}
\caption{Number of benchmarks completed by {\pbcount} when employing different compilation strategies, with 3600s timeout, higher number indicates better performance.}
\label{app:tab:coeff-modification-auction-result}
\end{table}

The `same value' row corresponds to setting (a), and we were able to get 887 non-zero count benchmarks from the original auction benchmarks. Similarly, setting (b) corresponds to the `3 values' row and we were able to get 903 non-zero count benchmarks. Notice that as we increase the number of different coefficients from setting (a) to setting (b), the number of benchmarks that can be completed by bottom-up compilation decreased from 758 to 705 even when the total number of benchmarks increased from 887 to 903. We also see that the gap between top-down and bottom-up approaches is reduced with the increase in different coefficient values.

\end{document}